\documentclass[pmlr,twocolumn,10pt]{jmlr} 

\usepackage{booktabs}
\usepackage{siunitx}

\usepackage{xcolor}         
\usepackage{graphicx}
\usepackage{amsmath}
\usepackage{caption}
\usepackage{multirow}
\usepackage{algpseudocode}
\usepackage{wrapfig}

\theorembodyfont{\upshape}
\theoremheaderfont{\scshape}
\theorempostheader{:}
\theoremsep{\newline}

\jmlrvolume{LEAVE UNSET}
\jmlryear{2023}
\jmlrsubmitted{LEAVE UNSET}
\jmlrpublished{LEAVE UNSET}
\jmlrworkshop{Conference on Health, Inference, and Learning (CHIL) 2023} 

\newcommand{\rebuttal}[1]{\textcolor{black}{#1}}

\title[MultiWave]{MultiWave: Multiresolution Deep Architectures through Wavelet Decomposition for Multivariate Time Series Prediction}

\author{%
\Name{Iman Deznabi}\Email{iman@cs.umass.edu}\\
\Name{Madalina Fiterau} \Email{mfiterau@cs.umass.edu}\\
\addr Manning College of Information \& Computer Sciences, University of Massachusetts Amherst, MA, USA
}

\begin{document}

\maketitle

\begin{abstract}
The analysis of multivariate time series data is challenging due to the various frequencies of signal changes that can occur over both short and long terms. Furthermore, standard deep learning models are often unsuitable for such datasets, as signals are typically sampled at different rates. To address these issues, we introduce MultiWave, a novel framework that enhances deep learning time series models by incorporating components that operate at the intrinsic frequencies of signals. MultiWave uses wavelets to decompose each signal into subsignals of varying frequencies and groups them into frequency bands. Each frequency band is handled by a different component of our model. A gating mechanism combines the output of the components to produce sparse models that use only specific signals at specific frequencies. Our experiments demonstrate that MultiWave accurately identifies informative frequency bands and improves the performance of various deep learning models, including LSTM, Transformer, and CNN-based models, for a wide range of applications. It attains top performance in stress and affect detection from wearables. It also increases the AUC of the best-performing model by 5\% for in-hospital COVID-19 mortality prediction from patient blood samples and for human activity recognition from accelerometer and gyroscope data. We show that MultiWave consistently identifies critical features and their frequency components, thus providing valuable insights into the applications studied.
\end{abstract}

\vspace{-0.5em}
\paragraph*{Data and Code Availability}
We are evaluating our models on three publicly available datasets, the Wearable Stress and Affect Detection (WESAD) \citep{schmidt2018introducing}, and MHEALTH dataset \citep{banos2014mhealthdroid, banos2015design} which both can be downloaded from the UCI Machine Learning Repository, and the COVID-19 dataset \citep{yan2020interpretable}. The code is available in \href{https://github.com/Information-Fusion-Lab-Umass/MultiWave}{https://github.com/Information-Fusion-Lab-Umass/MultiWave}.

\paragraph*{Institutional Review Board (IRB)}
This research does not require IRB approval.

\section{Introduction}
\label{sec:intro}

Multivariate time series prediction and forecasting are critical tasks in healthcare, as they enable the analysis and prediction of patient outcomes and the allocation of resources based on historical data. These applications play a key role in improving patient care and outcomes and are increasingly important in a healthcare industry that is becoming more data-driven \citep{wiens2018machine}.
However, the final prediction in these applications can depend on many factors, such as information at different frequencies, as well as long-term and short-term changes in input signals. Moreover, in many tasks, observations come from multiple sources and are often collected at various sampling rates. Here, we propose \emph{a model-agnostic approach that can leverage temporal dependencies at different frequencies and scales} in multivariate time series data. This approach is adept at handling multirate time series data, which refers to data collected at multiple sampling rates, through the use of multilevel discrete wavelet decomposition.

We have focused on three different applications of multivariate and multirate time series prediction: (1) stress and affect prediction using wearable devices, (2) COVID-19 in-hospital mortality prediction from blood samples collected over time, and (3) human activity recognition using wearable devices. The dataset corresponding to the first application is the Wearable Stress and Affect Detection (WESAD) dataset \citep{schmidt2018introducing}. This dataset includes physiological response data collected from a chest and a wrist-worn device from 15 subjects who were exposed to a baseline session, an amusement session, and a stress session. Our goal was to predict to which session the subjects were exposed on the basis of the physiological data collected. For COVID-19 in-hospital mortality prediction, we used the dataset provided by \cite{yan2020interpretable}. This dataset includes 74 biomarkers collected from 375 patient blood samples between January 10 and February 18, 2020, at Tongji Hospital in Wuhan, China. Our objective was to predict in-hospital mortality based on the collected time series of biomarkers. For human activity recognition, we used the Mobile Health (MHEALTH) dataset \citep{banos2014mhealthdroid, banos2015design}, which contains wearable sensor data from 10 volunteers performing 12 different physical activities.

Deep learning-based methods that are introduced into time series analysis, such as Recurrent Neural Networks (RNN) \citep{williams1989learning}, Convolutional Neural Networks (CNN) \citep{zheng2016exploiting} and more recently transformers \citep{wen2022transformers} achieve state-of-the-art results in these and similar applications \citep{lai2022lstm, tipirneni2021self, huang2022spatial, sun2021te, Ditch}. However, they have three notable shortcomings in handling multivariate time series data. First, these methods utilize only the information in the time domain and overlook the information available in the frequency domain of the signals. Second, they lack transparency in identifying the critical signals and subsignals that are relevant to the task at hand. Finally, they are unable to process signals that have been collected with varying frequencies (multirate signals) in their original form, which can lead to the introduction or loss of important temporal dependencies during up-sampling or down-sampling to a single rate \citep{che2018recurrent, tipirneni2021self, che2018hierarchical}.
We propose a new, model-agnostic approach to address these shortcomings. This framework employs discrete wavelet decomposition to segment signals into various frequency components (\emph{subsignals}). After eliminating irrelevant subsignals, it organizes subsignals with similar frequencies into individual time series models. Finally, the output from these models is combined to generate a prediction.
This framework brings the following improvements to multivariate time series modeling:
1) Uses the information available in both the time and frequency domains.
2) Provides unique insight into which frequencies of the signals are important for a given task.
3) Makes model-agnostic improvements to time series models.\\

Our results demonstrate that MultiWave enhances the performance of multivariate time series prediction models in wearable stress and affect prediction, COVID-19 mortality prediction, and human activity recognition. Moreover, MultiWave's capability to reveal useful signals and frequency components results in improved adaptability of deep learning models in these applications.

\section{Related work}
\begin{figure*}[t!]
    \centering
    \includegraphics[width=0.45\textwidth]{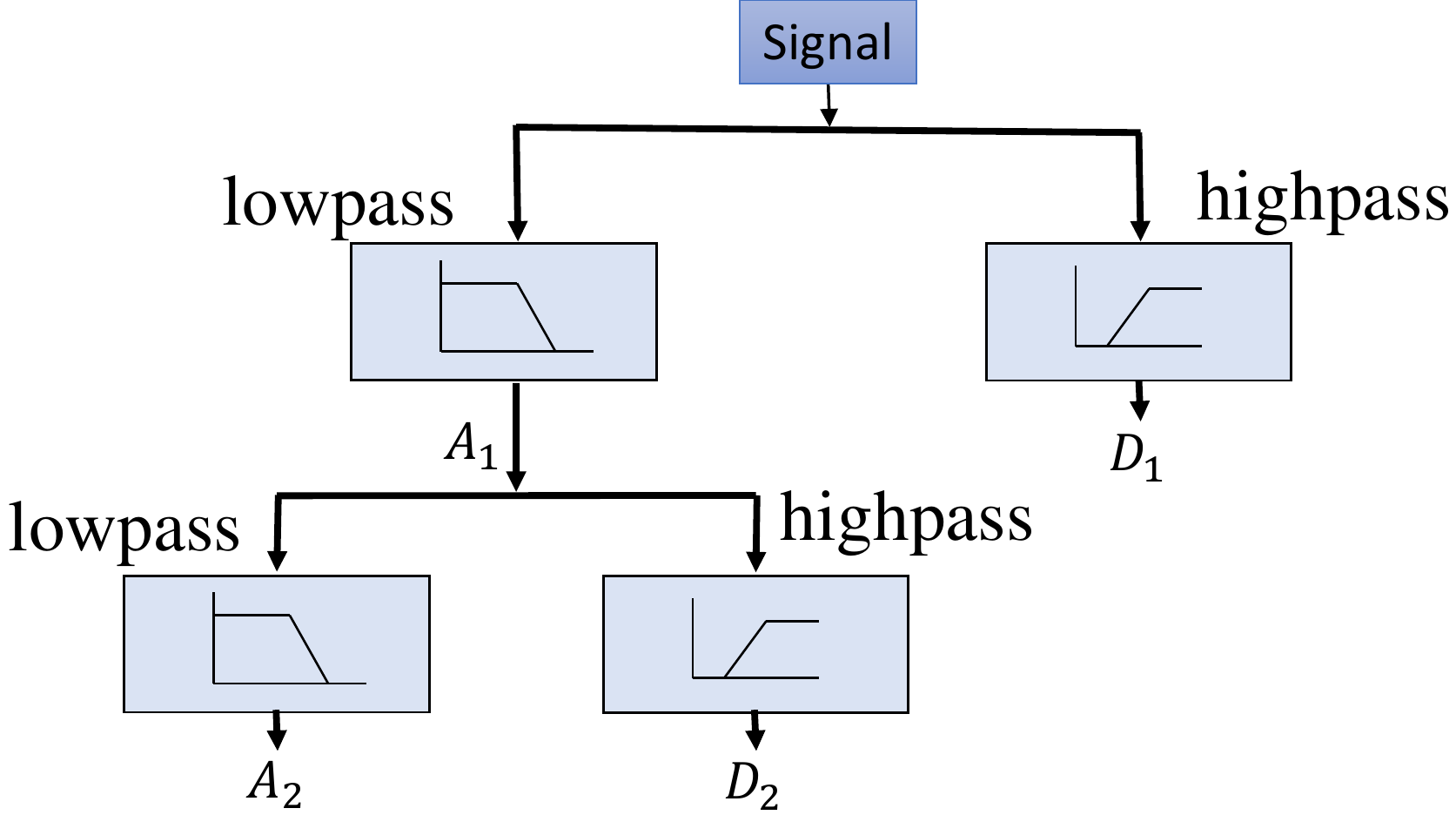}
    \includegraphics[width=0.4\textwidth]{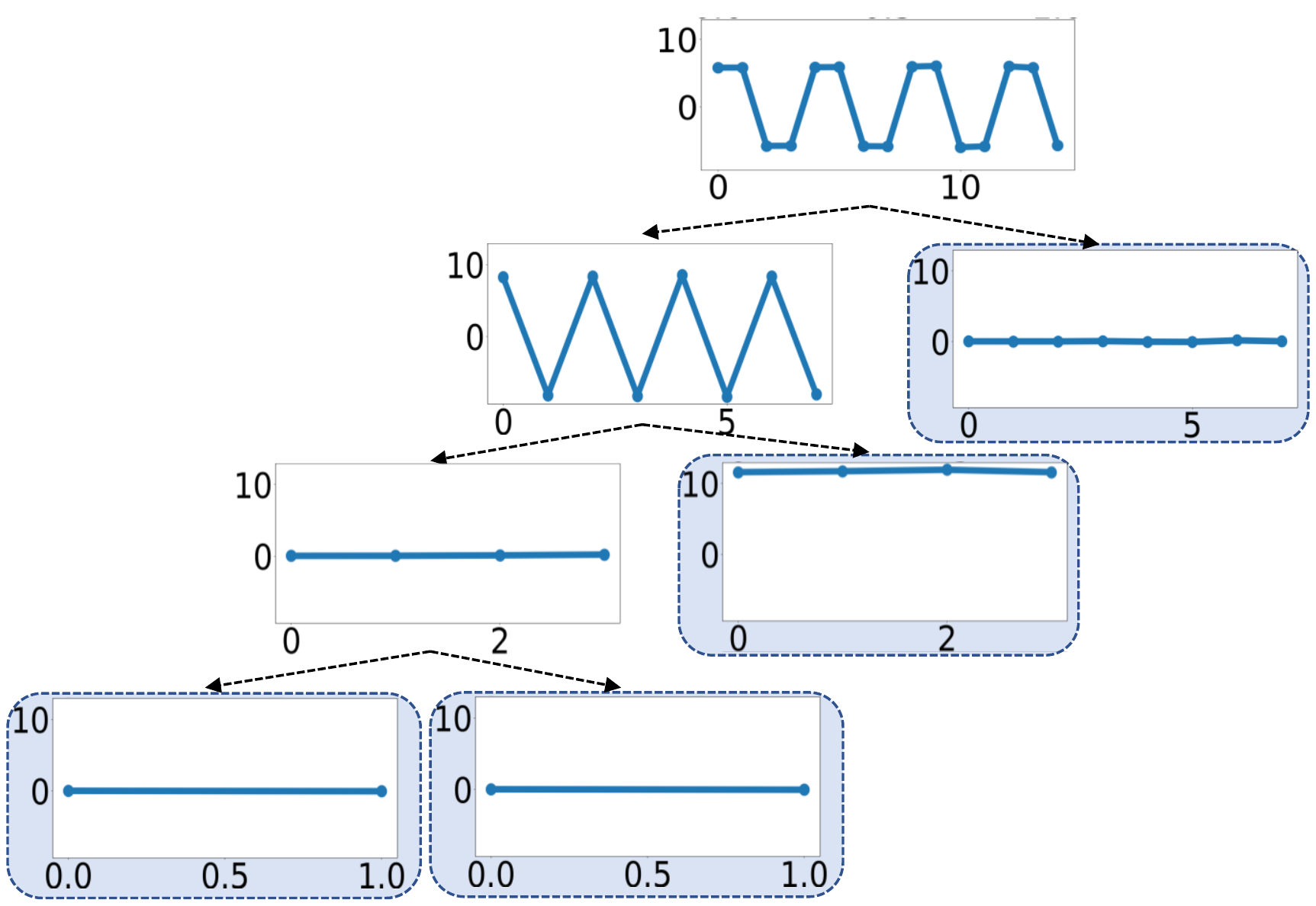}
    \caption{
     The process of multi-level discrete wavelet decomposition which uses lowpass and highpass filters to decompose signals. The image on the left depicts this process, while the image on the right showcases the decomposition of a signal using the Haar wavelet. As observed, the resulting signals demonstrate a value of zero, except for the signal component that corresponds to the true frequency of the original signal.}
    \label{fig:MDWD}
\end{figure*}
\textbf{Multirate time series classification}
Methods in time series analysis that are based on deep learning, such as Recurrent Neural Networks (RNN) \citep{williams1989learning}, Convolutional Neural Networks (CNN) \citep{zheng2016exploiting}, and more recently, Transformers \citep{wen2022transformers}, have demonstrated state-of-the-art results in numerous applications \citep{lai2022lstm, huang2022spatial}.
However, as mentioned earlier, they fall short when handling multirate time series data. There are several approaches proposed to solve this problem, such as \cite{che2018recurrent, tipirneni2021self} and \cite{chang2023tdstf}, which can inherently handle irregularly sampled time series data and thus can model multirate time series data without aligning the signals \citep{sun2020review, shukla2020survey}. There are other methods, such as \cite{che2018hierarchical, armesto2008multi, safari2014multirate} that use architectures specifically developed for multirate time series data. All these approaches only consider the information available in the time domain of the time series data, while MultiWave is specifically developed for multirate data and is able to leverage the information in the frequency domain.

\textbf{Frequency analysis of time series} Frequency analysis of time series is an extensively studied subject in the signal processing community. Methods such as the discrete Fourier transform \citep{bracewell1986fourier}, the discrete wavelet transform \citep{daubechies1992ten}, and the Z-Transform \citep{foster1996wavelets} have been used to analyze time series. For deep learning models, similar methods have been used in the preprocessing steps \citep{cui2016multi, yuan2017multi, song2021forecasting, salim2023comparative} or as part of neural networks \citep{koutnik2014clockwork, lee2021fnet}. Most of these models focus on univariate time series data and cannot be used directly on multivariate and multirate time series data.

\textbf{Wavelet decomposition} Wavelet decompositions \citep{daubechies1992ten} are well-known methods for capturing information in time series in both the time and frequency domains. They have been used successfully as a preprocessing step for neural networks \citep{LIU2013191, wang2020deep, alhnaity2021autoencoder, kim2021forecasting, althelaya2021combining, zucatelli2021investigation} and as an integral part of them \citep{SUBASI2006360, zhang1995wavelet, wang2018multilevel, guo2022hyperspectral, li2021hybrid, zhuang2022waveformer, zhou2022fedformer}. \citet{wang2018multilevel} proposes the methodology closest to our approach, implementing a trainable wavelet decomposition framework, which can be trained with the rest of the network. Although similar to our method, this paper uses wavelet decomposition to extract frequency-based information from datasets and \rebuttal{models} them using different model components; it is not suitable for utilization with multivariate and multirate time series data and does not incorporate the use of feature masks to eliminate frequency components that are not useful within the framework. Furthermore, their approach suggests a specific model to be employed, whereas our methodology is model-independent and can be integrated with any time series model.
\section{Background}
\subsection{Notation}
We denote multivariate and multirate time series data with $m$ signals collected before time $T$ as a set of signals $X^{1:T} = \{x^{1:T}_1, x^{1:T}_2, ..., x^{1:T}_m\}$ where each signal is collected at sampling rates $\{f_s(x_1), f_s(x_2), ..., f_s(x_m)\}$. The length of each signal is proportional to its collected rate $Len_i = T \times f_s(x_i)$. Given $X^{1:T}$, we want to predict a label $y$, which can be continuous (regression) or discrete (classification). In the rest of the paper, we will remove the time indication for the signals and show the set of signals as $X$ and the signal $i$ as $x_i$.
\subsection{Multilevel Discrete Wavelet Decomposition}
We use wavelet decomposition to break down the signals into different frequencies. Multilevel discrete wavelet decomposition can extract multilevel time-frequency features from a time series by iteratively applying low-pass and high-pass filters derived from wavelets to the signal. The formula for this decomposition is given below:
\begin{align*}
    x(t) =& \sum_k A_{L,k} \phi_{L,k}(t) + \sum_k D_{L,k}\Psi_{L,k}(t) + \\
                &\sum_k D_{L-1,k}\Psi_{L-1,k}(t) + ... + \sum_k D_{1,k}\Psi_{1,k}(t)
\end{align*}
$\Psi_{s, \tau}$ is the mother wavelet with scale $s$ and time $\tau$ and $\phi$ is the father wavelet. This multilevel wavelet decomposition converts the input signal $x(t)$ into signals $A_{L} = \bigcup_k A_{L,k}$, which is a coarse general approximation of the signal (low frequency) and the detail coefficients $D_{L} = \bigcup_k D_{L, k}, D_{L-1} = \bigcup_k D_{L-1, k}, \dots, D_{1} = \bigcup_k D_{1, k}$ that influence the function on various scales. Figure~\ref{fig:MDWD} depicts this decomposition. To simplify the notation, we show the decomposition of a signal $x$ as a set $S(x) = \{D_1, D_2, \dots, D_L, A_L\}$ that includes signals retrieved when decomposing $x(t)$ at $L$ levels. This notation allows us to denote $D_1$ as $S_1(x)$, $D_2$ as $S_2(x)$, \dots, $D_L$ as $S_L(x)$, and $A_L$ as $S_{L+1}(x)$.

Many different wavelets are introduced in the literature, such as Haar, Daubechies, and Biorthogonal \citep{haar1909theorie, cohen1992biorthogonal, daubechies1992ten}. Our framework is independent of the type of wavelet used. 

\begin{figure*}[t!]
    \centering
    \includegraphics[width=0.75\textwidth]{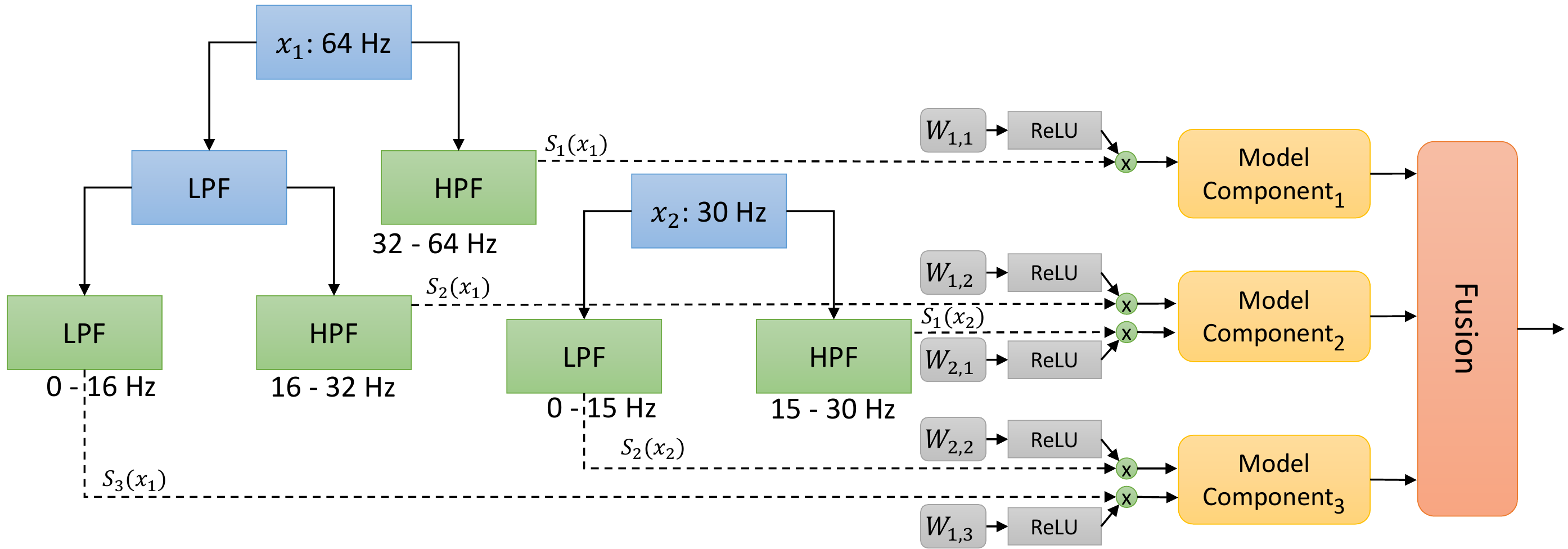}
    \caption{The structure of MultiWave with two signal inputs: one sampled at 64Hz ($x_1$) and the other at \rebuttal{30Hz} ($x_2$). Utilizing wavelet decomposition, the signals are decomposed into their corresponding frequency components using low-pass filters (LPF) and high-pass filters (HPF), which are subsequently grouped into distinct model components. Notably, this architecture is model-agnostic, thereby enabling its applicability to any model designed to process multivariate time series data.}
    \label{fig:model}
\end{figure*}
\section{MultiWave}
\subsection{Signal Decomposition and Frequency Grouping}
The overall structure of the framework is illustrated in Figure~\ref{fig:model} for the case where the signals $x_1$ and $x_2$ are collected at frequencies of $64Hz$ and $30Hz$, respectively. The framework uses discrete wavelet decomposition to decompose each signal into different frequency groups. For a set of $m$ input signals, $X = \{x_1, x_2, \dots, x_m\}$, a set of decomposed signals $\mathcal{S}(X) = \{S(x_1), S(x_2), \dots, S(x_m)\}$, will be generated. If all signals are sampled at the same rate, all elements of $\mathcal{S}(X)$ will have the same number of levels (denoted as $L$). In this case, all signals will have the same frequency at each level of decomposition; thus, all frequency components at the same level are grouped into a model component, denoted $\Phi_j$. The input of the component $j$ is $I_j = \{S_j(x_1), \dots, S_j(x_m)\}$ and the number of components is equal to the level of wavelet decomposition $L$. The outputs of all models are concatenated in the end and passed through a fully connected layer to obtain the final output.
\begin{equation}
    \hat{y} = FC(\Phi_1(I_1) \oplus \Phi_2(I_2) \oplus \dots \oplus \Phi_L(I_L))
\end{equation}

\begin{figure}
    \centering
    \includegraphics[width=0.3\textwidth]{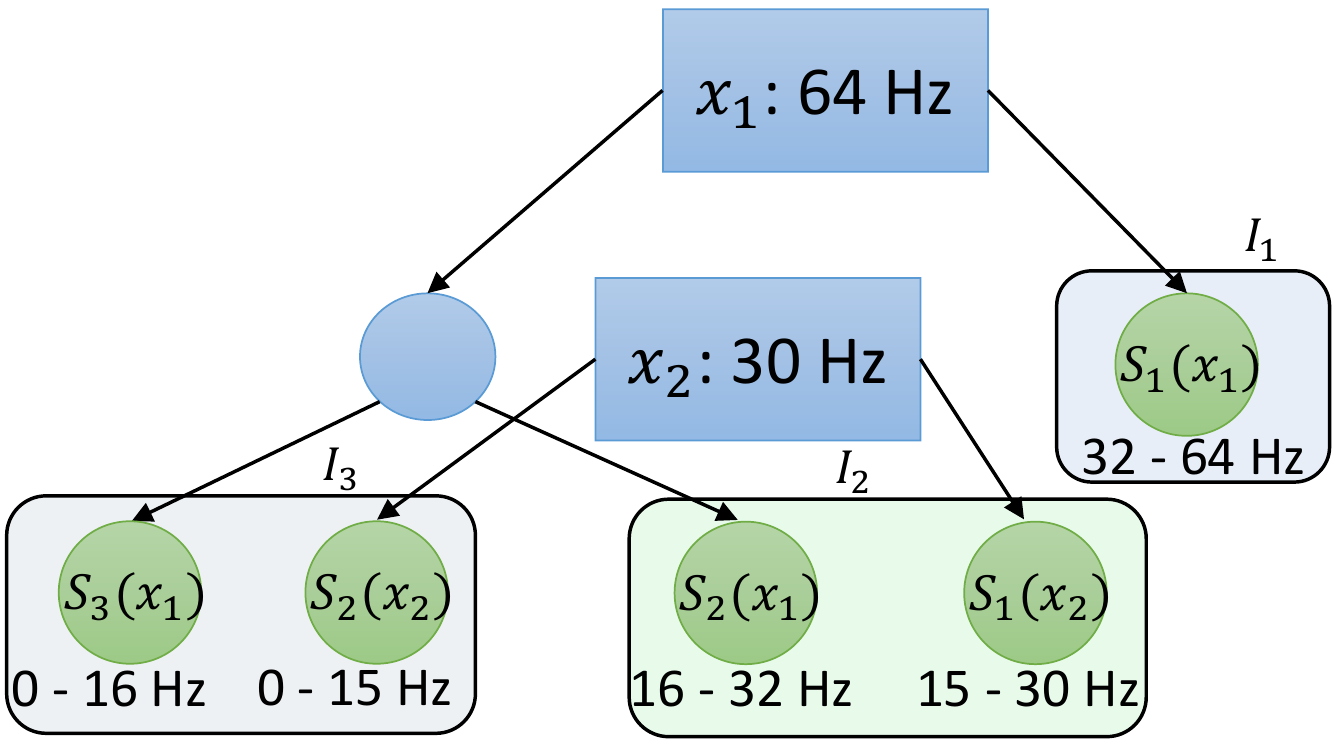}
    \caption{The decomposition and grouping of two signals sampled at 64Hz ($x_1$) and the other one at 30Hz ($x_2$). Three subsignals are extracted from $x_1$ through two levels of discrete wavelet decomposition, while $x_2$ is decomposed into two components in a single level. The subsignals are then organized into groups based on their frequency ranges.}
    \label{fig:dec_grouping}
\end{figure}

Here, $\oplus$ shows the concatenation operation. If the signals are collected with different sampling rates, we group the subsignals with frequencies that are close to each other together into a model component.
 In this case, the number of model components is equal to the maximum level of signal decomposition $L_{max} = max(\{L_1, L_2, \dots, L_m\})$, which is determined by the signal with the highest sampling rate. Here, $L_i$ represents the decomposition level for signal $i$. To better illustrate this, assume, without loss of generality, that we have two signals $m=2$ where signal $x_2$ is collected twice as often as signal $x_1$, $f_s(x_2) = 2 \times f_s(x_1)$. In this case, $L_2 = L_1 + 1$. Since the frequencies are reduced by half at each level of wavelet decomposition, we have: $$ f_s(S_{j}(x_1)) = f_s(S_{j+1}(x_2)) $$
To ensure that the signals have consistent rates in each component, the input for the first component of the model is defined as $I_1 = \{S_1(x_2)\}$. The input for the subsequent components, $j$, is expressed as: $I_j = \{S_{j-1}(x_1), S_j(x_2)\}$.
The difference between the number of levels between the decomposition of two signals is $\log({\frac{f_s(x_2)}{f_s(x_1)}})$, so if this proportion is not a power of $2$, the shorter signal at each component level, that receives both signals, is oversampled to the closest power of 2 to match the corresponding level of the other signal. For instance, if $f_s(x_2) = 6 \times f_s(x_1)$ in the above example, the input will be:
\begin{align*}
    I_1 &= \{S_1(x_2)\}, I_2 = \{S_2(x_2)\}, I_3 = \{S_3(x_2), S_1(x_1)\}\\ &\dots, I_{L_2} = \{S_{L_1}(x_1), S_{L_2}(x_2)\}
\end{align*}
since, for the component $\Phi_3$, $\frac{f_s(S_3(x_2))}{f_s(S_1(x_1))} = 1.5$, $S_1(x_1)$ is oversampled by a proportion of $1.5$. Note that the oversampling proportion will always be less than 2. Although this oversampling may lead to the same issues mentioned earlier, using the proposed method will considerably decrease the degree of oversampling, and the issues mentioned earlier would not occur to the same extent. Figure~\ref{fig:dec_grouping} shows how the decomposition and grouping procedures work for two signals with different sampling rates.

Using this approach, the components trained on lower frequencies learn long-term changes in the data, while the faster frequency components learn short-term fast-changing trends in the data. Furthermore, since the signals are grouped with respect to their sampling rates, the signals that are input into each component have similar frequencies, which significantly reduces the amount of oversampling.

\subsection{Masking frequency components}
Not all the frequency components of all signals are important for the prediction of the final label. To filter these frequencies, we introduce a learnable mask for each frequency component of the signals. We use ReLU activations \citep{nair2010rectified} on the weights of the model to mask the non-informative components by setting the mask to zero so the input of the component $i$ in the model is defined as:
\begin{align*}
    I_j = &\{ReLU{(W^{(1)}_j)} S_j(x_1), ReLU{(W^{(2)}_j)} S_j(x_2), \\ &..., ReLU{(W^{(m)}_j)} S_j(x_m)\}
\end{align*}
To force the model to use a sparse mask for the feature components, we add the $\ell_1$ norm of the weights to the final loss of the model:
$$ Loss = \mathcal{L} + \alpha \ell_1(ReLU{(W)})$$
where $\mathcal{L}$ is the normal loss of the model, which is usually defined by Mean-Squared-Error (MSE) for regression models and Cross-Entropy-Loss for classification models. $\alpha$ is a hyperparameter that determines the weight of the regularization term and sets a trade-off between minimizing the mask weights and the loss of the base model.

\subsection{Final fusion of components}
\label{sec:finalfusion}
We conducted experiments using various fusion techniques such as attention, mean, weighted average, ensemble methods, transformer fusion, and others like GradBlend \citep{wang2020makes} and Efficient Low-rank Multimodal Fusion with Modality-Specific Factors \citep{liu2018efficient, du2018adapting}. However, we discovered that concatenating and using a fully connected layer produced the best results. This may be because of the unique structure of the input data, where each input provides relevant information for the target and needs to be combined to produce the output. Additionally, each input's contribution to the final prediction may not be dependent on the other inputs. It is worth noting that MultiWave is not limited to any particular fusion technique, and different methods may work better for different datasets.


\subsection{Model training}

In our experiments, we used the Adam optimizer \citep{kingma2014adam} to train our models. Additionally, the feature masks of the baseline model, which takes in the original signals $X^{1:T} = {x^{1:T}_1, x^{1:T}_2, ..., x^{1:T}_m}$ as input, were incorporated as a supplementary component in the fusion model. This allows MultiWave to revert to the baseline model (using a form of early fusion for frequencies) if the incorporation of frequency components does not lead to improvement, while still having the option to utilize frequency components for enhanced performance. We also used weights and biases \citep{wandb} to track and record the experiments, and Pytorch \citep{NEURIPS2019_9015} to implement and train our models.


\section{Experiments}
\begin{figure*}[t!]
	\begin{center}
		\begin{tabular}{c c}
			{\includegraphics[width=0.4\textwidth]{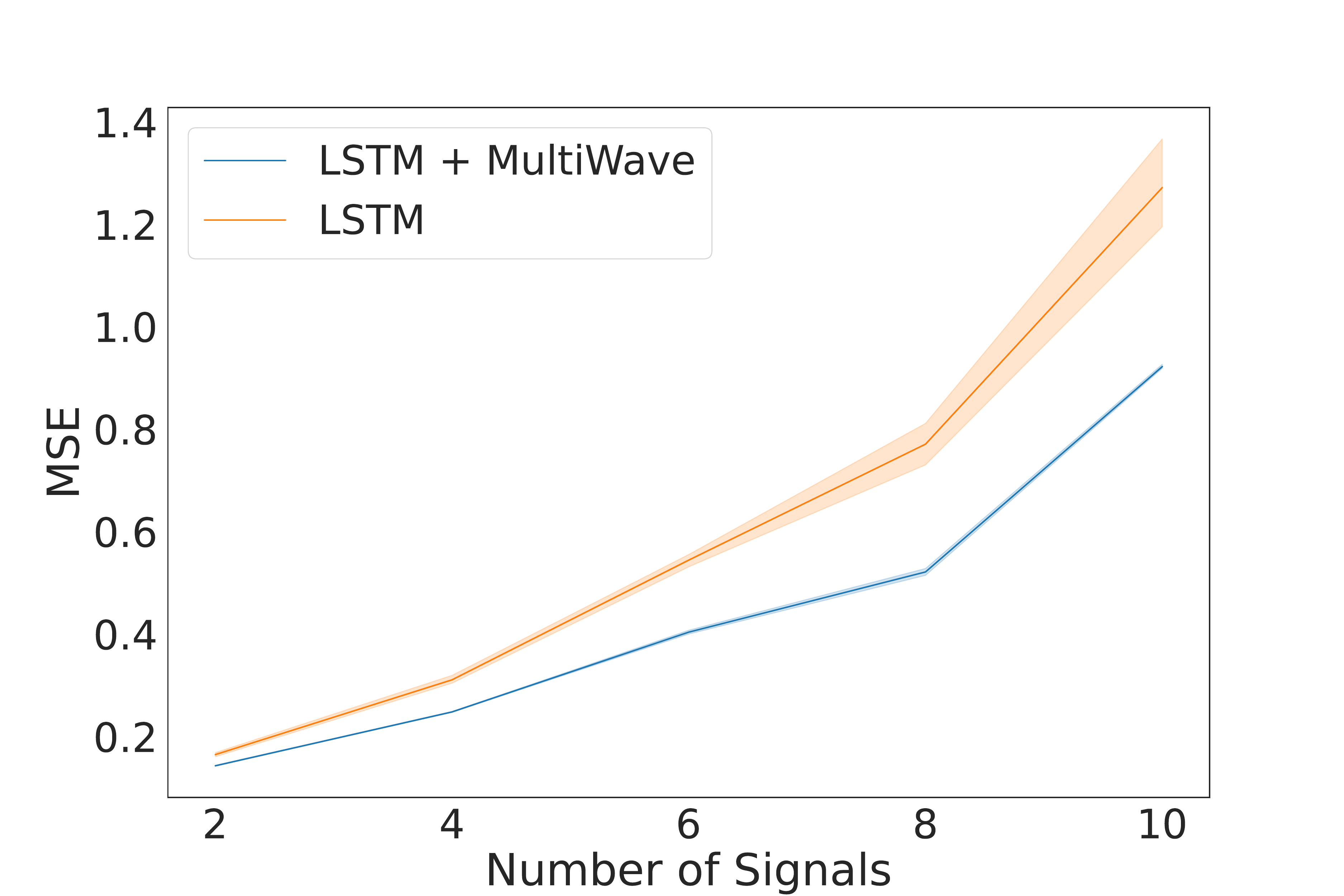}} & \includegraphics[width=0.4\textwidth]{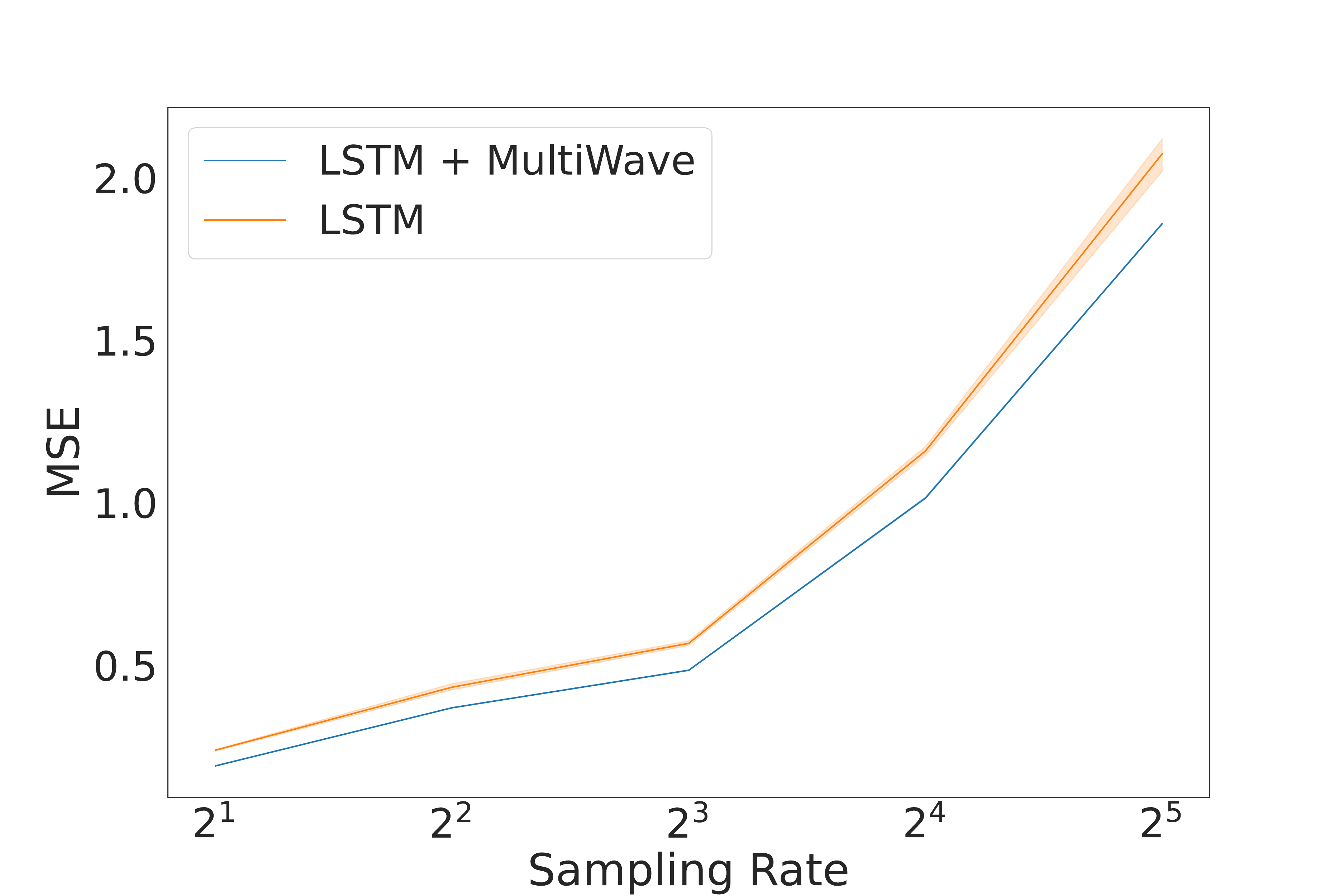}\\
			{\includegraphics[width=0.4\textwidth]{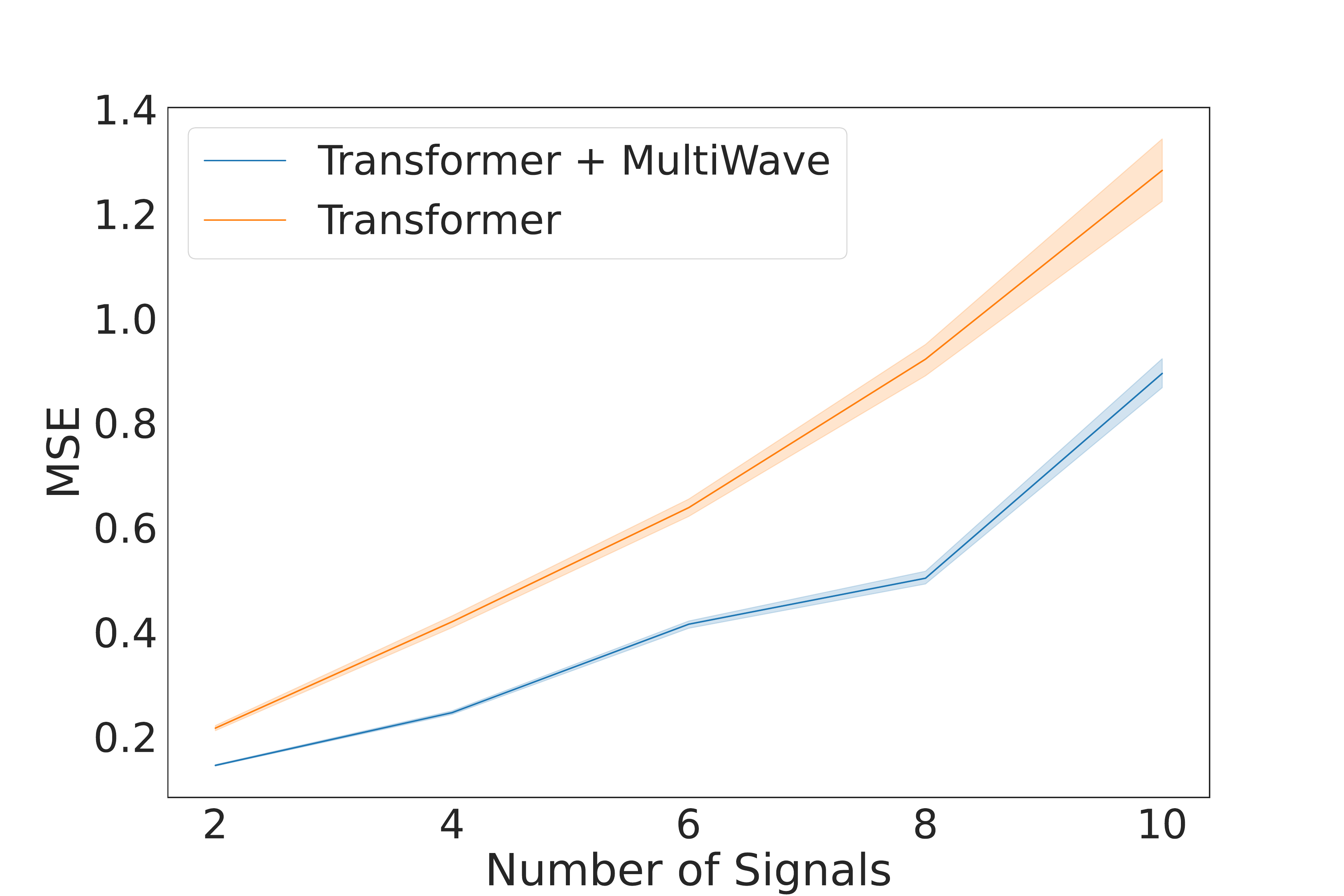}} & \includegraphics[width=0.4\textwidth]{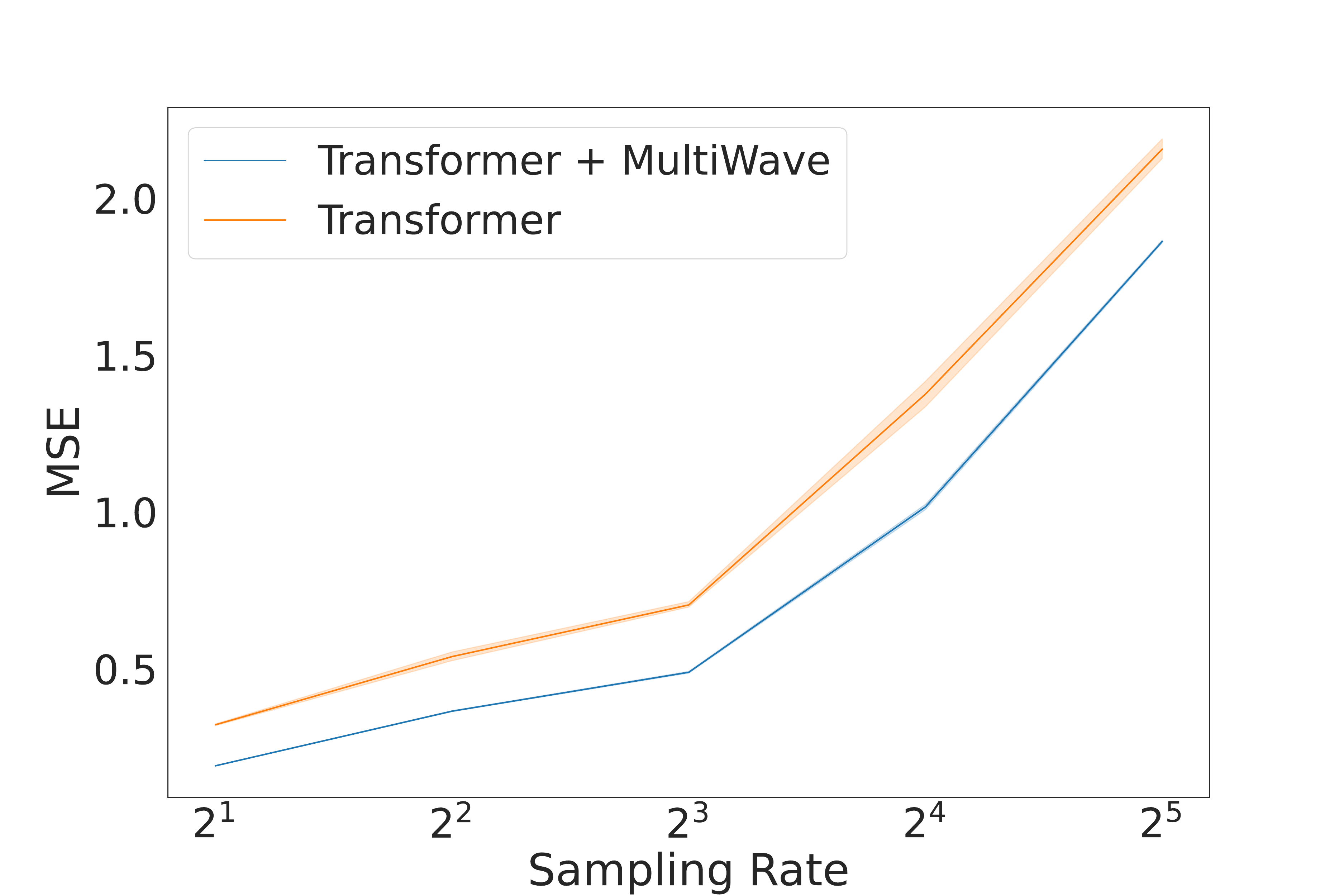}
		\end{tabular}
		\caption{The Mean Squared Error (MSE) results for a synthetic dataset, with two columns displaying the effects of varying the number of signals and the difference between the signal sampling rates, respectively. The first row of the figure showcases the results for the LSTM model, while the second row illustrates the results for Transformers. Notably, the utilization of Multiwave results in improved performance of the models. This improvement can be attributed to Multiwave's ability to identify and filter out irrelevant frequency components of the signals, consequently reducing the need for upsampling and mitigating errors caused by imputation.}
		\label{fig:SynResults}
	\end{center}
\end{figure*}
In this section, we evaluate the performance of MultiWave in synthetically generated and real-world datasets. \rebuttal{We used Haar wavelet functions \citep{haar1909theorie} throughout the experiments and, as mentioned in Section~\ref{sec:finalfusion}, we used a concatenation and a fully connected layer for the fusion of the output of model components.}
\subsection{Synthetically generated data}
\label{sec:syn_data}
To determine the effectiveness of MultiWave in handling signals with different frequencies and sampling rates, we generated synthetic data. The generated data consist of multiple square signals, each with a different frequency and amplitude. The amplitude of each signal is randomly selected from a uniform distribution ranging from 0-10. We then add uniform noise to the input data with an amplitude of 3.
The label $y$ is the sum of the amplitudes of the generated signals \rebuttal{before the noise is added}. The task is to predict the label $y$ from given time series data. Figure~\ref{fig:SynResults} shows the results of the two experiments. \rebuttal{For each experiment, we ran each setting five times with different random seeds and displayed the mean and standard deviation of these experiments}. 
\rebuttal{In the first experiment, two square wave signals were used as input with frequencies of 1Hz and 2Hz, sampled at a rate of 128Hz for a duration of 1 second. Then, additional signals were iteratively added with frequencies of 4Hz, 4Hz, 8Hz, 8Hz, 16Hz, 16Hz, 32Hz, and 32Hz, respectively, all sampled at the same rate of 128Hz. The MultiWave model was used with both LSTM and Transformer components, and the results are shown in the first column of Figure}~\ref{fig:SynResults}\rebuttal{. The MultiWave model was able to improve performance steadily and reliably with the addition of new signals, owing to its ability to identify significant frequency components, eliminate noise, and make more precise predictions.}

\rebuttal{In the second experiment, two square wave signals were generated with frequencies of 2Hz and 4Hz, sampled at rates of 64Hz and 128Hz, respectively, for 1 second. The sampling rate of the first signal was then reduced to 32Hz, 16Hz, 8Hz, and 4Hz. The results for the LSTM and Transformer models, with and without the use of MultiWave, are shown in the second column of Figure}~\ref{fig:SynResults}\rebuttal{. The MultiWave model consistently improved the performance of the baseline models by grouping signals with similar sampling rates, which reduced the need for imputation and the potential for errors that can affect the accuracy of the model predictions.}

\rebuttal{Synthetic data experiments with frequencies other than the powers of two are in Appendix Section~\ref{sec:challenging_Syn}.}
\subsection{The Wearable Stress and Affect Detection (WESAD)}
\begin{table*}
    \centering
    \begin{tabular}{|c|c|c|c|}
        \hline
        Dataset & Model & AUC without MultiWave & AUC with MultiWave \\[0.2ex] \hline\hline
        {\multirow{6}{*}{WESAD}} & \rule{0pt}{2ex} LSTM & $0.822 \pm 0.04$ & $\mathbf{0.828 \pm 0.04}$ \\[0.2ex] \cline{2-4}
                                 & \rule{0pt}{2ex} CNN-Attn & $0.831 \pm 0.03$ & $\mathbf{0.877 \pm 0.03}$ \\[0.2ex] \cline{2-4}
                                 & \rule{0pt}{2ex} CNN-LSTM & $0.807 \pm 0.04$ & $\mathbf{0.839 \pm 0.04}$ \\[0.2ex] \cline{2-4}
                                 & \rule{0pt}{2ex} Transformer & $0.805 \pm 0.04$ & $\mathbf{0.824 \pm 0.03}$ \\[0.2ex] \cline{2-4}
                                 & \rule{0pt}{2ex} FCN & $0.805 \pm 0.04$ & $\mathbf{0.833 \pm 0.05}$\\[0.2ex] \cline{2-4}
                                 & \rule{0pt}{2ex} FCN-PC & $0.860 \pm 0.04$ & $\mathbf{0.904 \pm 0.03}$
                                 \\[0.2ex] \hline\hline
        {\multirow{4}{*}{\shortstack{COVID-19 \\ 0 days ahead}}} & \rule{0pt}{2ex} LSTM & $0.983 \pm 0.008$  & $\mathbf{0.989 \pm 0.004}$ \\[0.2ex] \cline{2-4}
                                    & \rule{0pt}{2ex} CNN-Attn & $0.978 \pm 0.012$ & $\mathbf{0.979 \pm 0.02}$ \\[0.2ex] \cline{2-4}
                                    & \rule{0pt}{2ex} CNN-LSTM & $0.979 \pm 0.010$ & $\mathbf{0.981 \pm 0.012}$ \\[0.2ex] \cline{2-4}
                                    & \rule{0pt}{2ex} Transformer & $0.980 \pm 0.007$ & $\mathbf{0.984 \pm 0.008}$ \\[0.2ex] \hline\hline
        {\multirow{4}{*}{\shortstack{COVID-19 \\ 12 days ahead}}} & \rule{0pt}{2ex} LSTM & $0.977 \pm 0.007$ & $\mathbf{0.979 \pm 0.006}$ \\[0.2ex] \cline{2-4}
                                    & \rule{0pt}{2ex} CNN-Attn & $\mathbf{0.967 \pm 0.008}$ & $\mathbf{0.967 \pm 0.013}$ \\[0.2ex] \cline{2-4}
                                    & \rule{0pt}{2ex} CNN-LSTM & $0.961 \pm 0.012$ & $\mathbf{0.962 \pm 0.009}$ \\[0.2ex] \cline{2-4}
                                    & \rule{0pt}{2ex} Transformer & $0.969 \pm 0.01$ & $\mathbf{0.972 \pm 0.01}$ \\[0.2ex] \hline
    \end{tabular}
    \caption{The AUC performance of MultiWave on real-world datasets WESAD and COVID-19, both with and without the use of MultiWave (when the model is used as $\Phi$ component in MultiWave architecture). CNN-Attn is a Convolutional Neural Network (CNN) model followed by an attention layer; CNN-LSTM is a CNN model followed by a Long Short-Term Memory (LSTM) layer; FCN is a fully convolutional neural network; and FCN-PC is a per-channel FCN model, as described in \cite{Ditch}.}
    \label{tab:RealResults}
\end{table*}

 Wearable Stress and Affect Detection (WESAD) \citep{schmidt2018introducing} is a publicly available multimodal dataset for stress and affect detection. WESAD contains physiological response data from 15 subjects during three sessions of baseline, amusement, and stress. The baseline session is 20 minutes, where the subject is doing a neutral reading task, the amusement session is watching a set of funny videos for 392 seconds, and the stress session is when the subject is exposed to the Trier Social Stress Test \citep{kirschbaum1993trier} for 10 minutes. During these sessions, various physiological measurements are taken using a chest-worn device and a wrist-worn device. Measurements include blood volume pulse (BVP), electrocardiogram (ECG), electrodermal activity (EDA), electromyogram (EMG), respiration (RESP), temperature (TEMP), and accelerometer (ACC). The chest-worn device collects data at a frequency of 700 Hz, while the wrist-worn device collects data at frequencies of 64 Hz, 32 Hz, and 4 Hz. We followed the data preprocessing method described in \cite{Ditch} but used sampling rates that are powers of 2 to obtain more consistent signals. More details on the preprocessing of the signals in this dataset are given in the Supplementary Section~\ref{sec:app_WESAD_preprocess}.

The results achieved in this dataset are shown in Table~\ref{tab:RealResults}. To our knowledge, the fully convolutional per channel neural network (FCN-PC) described in the paper \cite{Ditch} is the state-of-the-art deep learning model for this dataset in these prediction settings, and MultiWave brings more than 5\% improvement to this model. 
MultiWave improves the performance of all baseline models since it allows the model components to learn short-term and long-term changes and can combine these multirate signals without the need for alignment and imputation. Table~\ref{tab:WESAD_Masks} shows the signals and their frequency components consistently selected by the MultiWave model. The lower frequencies of the accelerometer data are selected, which is consistent with previous research \citep{10.1242/jeb.184085, rooney2022prediction, alhassan2010defining}. Other physiological data were also consistently selected at lower frequencies, which aligns with earlier research that highlights the significance of summary statistics for this type of data \citep{rooney2022prediction}. More details about these experiments are given in the Supplementary Section~\ref{sec:app_WESAD}.
\rebuttal{Additionally, to investigate whether the performance improvements of the models are solely attributable to the inclusion of frequency features, we constructed a naive model that utilizes both FFT features and the signal itself. The architecture of this model is illustrated in Appendix Section~\ref{sec:ModelFFT}. We used the same model component for both FFT features and the input signal and concatenated the results before passing them through a 2-layer fully connected network. The experimental results are presented in Table~\ref{tab:WESADFFT}.}

\begin{table*}
    \centering
    \begin{tabular}{|c|p{8cm}|}
        \hline
        Frequency component & Features \\ \hline
        \rule{0pt}{2.6ex} $0 - 4$ Hz \rule[-1.2ex]{0pt}{0pt} & Chest acceleration Y, Chest acceleration Z, Wrist acceleration X, Chest ECG, Wrist BVP, Wrist EDA, Wrist temperature  \\ \hline
        \rule{0pt}{2.6ex} $4 - 8$ Hz\rule[-1.2ex]{0pt}{0pt} & Chest ECG \\ \hline
    \end{tabular}
    \caption{features with non-zero mask weights after the training procedure on the WESAD dataset. These features have been consistently selected in 5 separate runs of the training, demonstrating their significance for the predictive task. This aligns with the domain expertise and supports the ability of MultiWave to automatically identify relevant feature dependencies, as well as determining the essential subsignals among different frequencies.}
    \label{tab:WESAD_Masks}
\end{table*}

\begin{table*}
\centering
        \begin{tabular}{|c|c|c|c|}
        \hline
        Component & AUC Simple & AUC w/ FFT & AUC w/ MultiWave \\\hline
        LSTM & $0.822 \pm 0.04$ & $0.816 \pm 0.03$ & $\mathbf{0.828 \pm 0.04}$ \\\hline
        CNN-Attn & $0.831 \pm 0.03$ & $0.816 \pm 0.05$ & $\mathbf{0.877 \pm 0.03}$ \\\hline
        CNN-LSTM & $0.807 \pm 0.04$ & $0.835 \pm 0.04$ & $\mathbf{0.839 \pm 0.04}$ \\\hline
        Transformer & $0.805 \pm 0.04$ & $0.813 \pm 0.05$ & $\mathbf{0.824 \pm 0.03}$ \\\hline
        FCN & $0.805 \pm 0.04$ & $0.800 \pm 0.05$ & $\mathbf{0.833 \pm 0.05}$\\\hline
        FCN-PC & $0.860 \pm 0.04$ & $0.884 \pm 0.04$ & $\mathbf{0.904 \pm 0.03}$ \\\hline
    \end{tabular}
    \caption{\rebuttal{This table compares the AUC results of models using FFT features versus those using MultiWave on the WESAD dataset. Some model components struggle to handle FFT values, and since we're adding complexity to the model, there may not be sufficient data for the model with FFT features to exclusively learn from time domain features. Additionally, because we're splitting the parameters between the two components, there are cases where adding FFT features results in worse performance than the baseline model.}}
    \label{tab:WESADFFT}
\end{table*}

\begin{table*}
    \centering
    \begin{tabular}{|c|c|}
        \hline
        Frequency component & Features \\ \hline
        \rule{0pt}{2.6ex} $0 - \frac{1}{16}$ days \rule[-1.2ex]{0pt}{0pt} & \textbf{High sensitivity C-reactive protein}, Glucose  \\ \hline
        \rule{0pt}{2.6ex} $\frac{1}{16} - \frac{1}{8}$ days \rule[-1.2ex]{0pt}{0pt} & \textbf{Lactate dehydrogenase} \\ \hline
        \rule{0pt}{2.6ex} $\frac{1}{8} - \frac{1}{4}$ days \rule[-1.2ex]{0pt}{0pt} & D-D dimer \\ \hline
        \rule{0pt}{2.6ex} $\frac{1}{4} - \frac{1}{2}$ days \rule[-1.2ex]{0pt}{0pt} & - \\ \hline
        \rule{0pt}{2.6ex} $\frac{1}{2} - \frac{1}{1}$ days \rule[-1.2ex]{0pt}{0pt} & \textbf{(\%)lymphocyte}, \textbf{High sensitivity C-reactive protein} \\ \hline
    \end{tabular}
    \caption{The features with nonzero mask weights in the COVID-19 dataset at the end of the training. These features were consistently selected in 5 different runs of the training procedure, indicating that they are informative for the prediction task, which is consistent with domain expertise. MultiWave was able to automatically determine these dependencies, as well as determine which subsignals of different frequencies are relevant, as shown in Figure~\ref{fig:COVID_Masks_Time}}
    \label{tab:COVID_Masks}
\end{table*}

\subsection{COVID-19}
The COVID-19 dataset \citep{yan2020interpretable} is a publicly available dataset that contains 74 indicators of 375 patient blood samples from 10 January to 18 February 2020 at Tongji Hospital, Wuhan, China. These indicators are collected in irregular time intervals, and sampling rates range from 0 to 6 per day. We are interested in the task of predicting in-hospital mortality given the time series of biomarkers. The duration of hospital stays for patients varies from 2 hours to 35 days.

\begin{figure*}[!ht]
    \centering
    \includegraphics[width=0.85\textwidth]{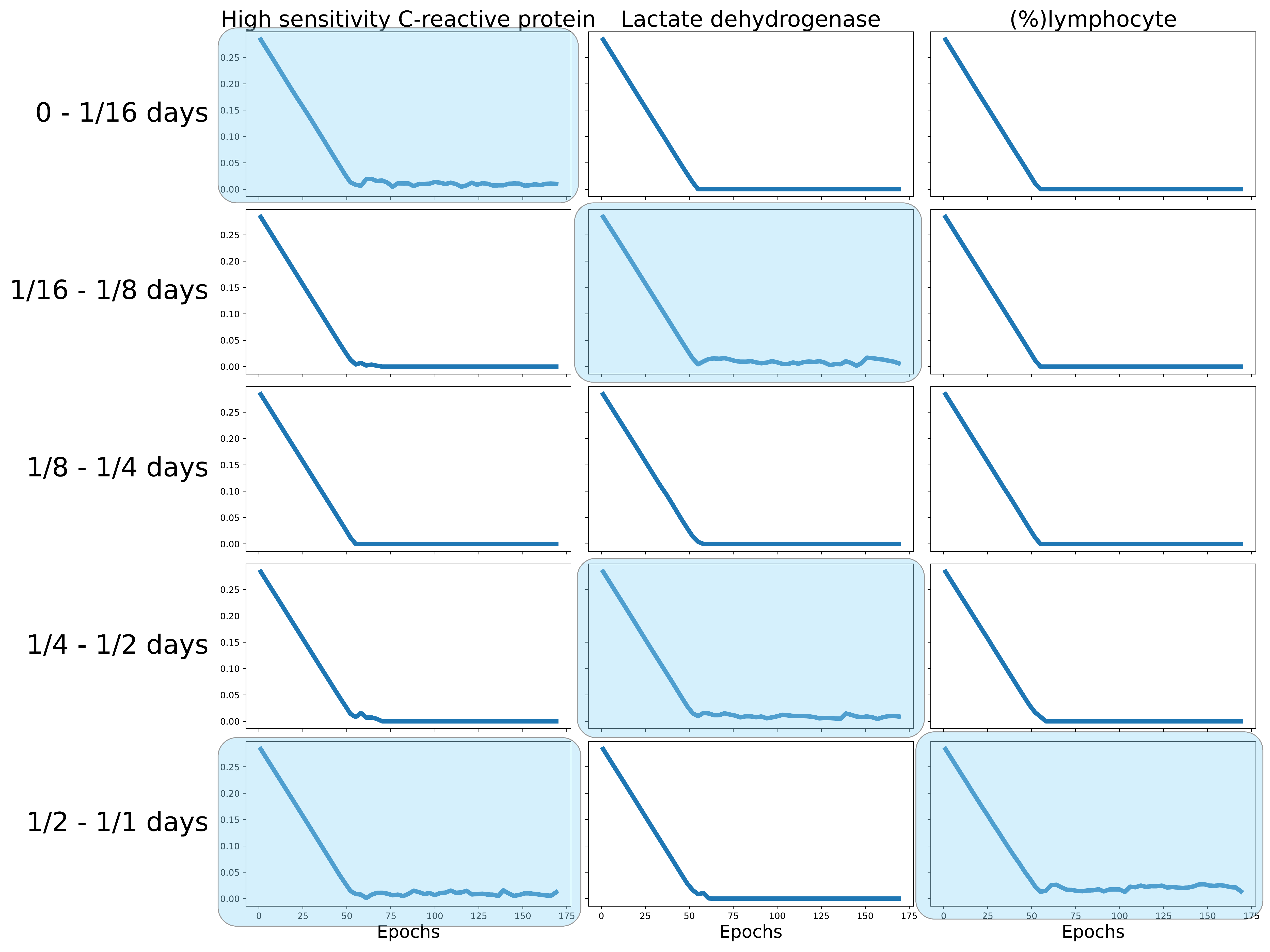}
    \caption{The mask weights over training epochs for the three most important features for different frequency components in COVID-19 dataset. Masks that are not zeroed out are highlighted.}
    \label{fig:COVID_Masks_Time}
\end{figure*}

 To process this dataset, we sampled the features with different rates ranging from 1 to 8-day intervals (more details in the Supplementary Section~\ref{sec:app_COVID}). If multiple values are recorded for a feature in the determined rate, we use the last recorded value. We fill in the missing values by linear interpolation. if a feature is completely missing for a patient, we impute it using that feature's mean value (i.e., the average of the recorded values for all patients in the training set). To evaluate the capability to select features in our model, we used all 74 features. Because of this, we were unable to use the original test set provided with the dataset as it only contains 3 features. Therefore, to evaluate our models, we separated 100 patients from the original dataset and used 50 patients for validation and 50 patients to test our models.

 The first group of experiments aims to predict mortality at the time of hospital discharge (0 days ahead of the prediction) by considering all patient data leading up to discharge. The second group of experiments is focused on predicting mortality 12 days ahead of hospital discharge (12 days ahead of prediction), incorporating patient data up to that point. The average AUC results for five runs of each experiment are included in Table~\ref{tab:RealResults}. For this dataset, we used the LSTM, CNN-Attn, CNN-LSTM, and Transformer models, and the results are reported with and without the inclusion of the MultiWave framework. Consistent with the results reported in \cite{sun2021te}, LSTM-based models achieve the best results, and MultiWave brings consistent improvements to all baseline models.
 
 \cite{yan2020interpretable} reports three features as the most important characteristics in the prediction of hospital mortality in this dataset: lactic dehydrogenase, lymphocytes, and high-sensitivity C-reactive protein. To determine whether MultiWave can recognize the correct features in predicting the target, we looked at non-zero masked values in each model component for five different runs, and we show the common ones in Table~\ref{tab:COVID_Masks}. As can be seen, the model consistently selects these three features as important predictors, and the model has identified important subsignals within these features. \rebuttal{These features are also reported to be important by \cite{gao2021risk}, \cite{sun2021te} and \cite{jiang2021peripheral}}. In Figure~\ref{fig:COVID_Masks_Time} we show how the mask weights for these features across the components are changed over the training epochs of the model for one run.

\subsection{MHEALTH}
The MHEALTH \citep{banos2014mhealthdroid,banos2015design} dataset is a publicly available dataset for health monitoring applications that contains data from 10 volunteers with diverse profiles while performing 12 different physical activities. The dataset was collected using Shimmer2 wearable sensors \citep{burns2010shimmer}, which were placed on the subject's chest, right wrist, and left ankle and attached using elastic straps. The dataset includes various physiological measurements, including acceleration, rate of turn, and magnetic field orientation, recorded at a sampling rate of 50 Hz. In addition, the chest sensor provides 2-lead ECG measurements.

In this dataset, we are trying to recognize different activities and classify them into 13 classes, which include 12 activities and the absence of any activity. To do this, we separated our dataset into two sets: a test set consisting of data from two subjects and a training set consisting of data from the other eight subjects. We only used acceleration and gyroscope data for this experiment (for more details, see Supplementary Section~\ref{sec:app_Mhealth}).

We conducted two different experiments on this dataset, one with the original uniformly sampled data, and for the second experiment, we resampled the features with frequencies of $1/3, 1/5, 1/7, ..., 1/25$ to obtain a multirate dataset. Because the classes are balanced, and following prior research \citep{demrozi2020human, chen2021deep} we report the classification accuracy in this dataset in Table~\ref{tab:Mhealth}. MultiWave brings consistent improvement to all baseline models, and the LSTM model with the inclusion of MultiWave achieved the best accuracy in the uniformly sampled dataset in which MultiWave brings an improvement of $9.8\%$.
In the multirate version of the dataset the CNN-LSTM model with MultiWave achieved the best accuracy in which MultiWave brought an improvement of $5.6\%$. Table~\ref{tab:Mhealth_Masks} shows the signals and their frequency components that were consistently selected by the MultiWave model. Accelerometer data are selected with different frequencies, consistent with previous research showing the importance of accelerometer data in activity recognition\citep{stisen2015smart,ordonez2016deep}.

\begin{table*}
    \centering
    \begin{tabular}{|c|c|c|c|}
        \hline
        Dataset & Model & Accuracy without MultiWave & Accuracy with MultiWave \\[0.2ex] \hline\hline
        {\multirow{4}{*}{\shortstack{MHEALTH}}} & \rule{0pt}{2ex} LSTM & $88.90\% \pm 4.48$ & $\mathbf{97.59\% \pm 0.68}$ \\[0.2ex] \cline{2-4}
                                    & \rule{0pt}{2ex} CNN-Attn & $92.90\% \pm 2.69$ & $\mathbf{95.50\% \pm 2.11}$ \\[0.2ex] \cline{2-4}
                                    & \rule{0pt}{2ex} CNN-LSTM & $92.04\% \pm 3.33$ & $\mathbf{96.48\% \pm 1.20}$ \\[0.2ex] \cline{2-4}
                                    & \rule{0pt}{2ex} Transformer & $95.83\% \pm 3.53$ & $\mathbf{96.14\% \pm 1.14}$ \\[0.2ex] \hline\hline
        {\multirow{4}{*}{\shortstack{MHEALTH \\ Multirate}}} & \rule{0pt}{2ex} LSTM & $81.84\% \pm 3.68$ & $\mathbf{92.34\% \pm 0.69}$ \\[0.2ex] \cline{2-4}
                                    & \rule{0pt}{2ex} CNN-Attn & $90.24\% \pm 2.70$ & $\mathbf{94.03\% \pm 1.34}$ \\[0.2ex] \cline{2-4}
                                    & \rule{0pt}{2ex} CNN-LSTM & $90.19\% \pm 3.94$ & $\mathbf{95.28\% \pm 0.41}$ \\[0.2ex] \cline{2-4}
                                    & \rule{0pt}{2ex} Transformer & $91.16\% \pm 3.86$ & $\mathbf{91.52\% \pm 1.63}$ \\[0.2ex] \hline
    \end{tabular}
    \caption{The accuracy performance of MultiWave on the MHEALTH dataset both with and without the use of MultiWave (when the model is used as $\Phi$ component in MultiWave architecture). The first part of the table shows the results for the original dataset (uniformly sampled data), and the second part shows the results when the dataset is resampled to get a multirate dataset.}
    \label{tab:Mhealth}
\end{table*}

\begin{table*}
    \centering
    \begin{tabular}{|c|p{8cm}|}
        \hline
        Frequency component & Features \\ \hline
        \rule{0pt}{2.6ex} $0 - 3.125$ Hz \rule[-1.2ex]{0pt}{0pt} & Left-ankle acceleration (X-axis, Y-axis, Z-axis), Right-lower-arm acceleration (X-axis, Y-axis, Z-axis) \\ \hline
        \rule{0pt}{2.6ex} $3.125 - 6.25$ Hz\rule[-1.2ex]{0pt}{0pt} & Left-ankle acceleration (Y-axis, Z-axis), Right-lower-arm acceleration (Y-axis) \\ \hline
        \rule{0pt}{2.6ex} $6.25 - 12.5$ Hz\rule[-1.2ex]{0pt}{0pt} & Left-ankle acceleration (Y-axis, Z-axis)\\ \hline
        \rule{0pt}{2.6ex} $12.5 - 25$ Hz\rule[-1.2ex]{0pt}{0pt} & Right-lower-arm acceleration (Y-axis) \\ \hline
        \rule{0pt}{2.6ex} $25 - 50$ Hz\rule[-1.2ex]{0pt}{0pt} & Left-ankle acceleration (X-axis), Right-lower-arm acceleration (Y-axis) \\ \hline
    \end{tabular}
    \caption{Features with nonzero masks after the training procedure on the MHEALTH dataset. These features have been consistently selected in 5 separate training runs, demonstrating their significance for the prediction task.}
    \label{tab:Mhealth_Masks}
\end{table*}

 \begin{table*}
\centering
        \begin{tabular}{|c|c|c|c|}
        \hline
        Dataset & AUC w/o MultiWave & AUC w/ MultiWave No Mask & AUC w/ MultiWave \\\hline
        WESAD & $0.831 \pm 0.03$ & $0.866 \pm 0.06$ & $\mathbf{0.877 \pm 0.03}$ \\\hline
        COVID-19 0 d ahead & $0.980 \pm 0.007$ & $\mathbf{0.984 \pm 0.01}$ & $\mathbf{0.984 \pm 0.008}$ \\\hline
        COVID-19 12 d ahead & $0.969 \pm 0.01$ & $0.964 \pm 0.02$ & $\mathbf{0.972 \pm 0.01}$ \\\hline\hline
         & Acc w/o MultiWave & Acc w/ MultiWave No Mask & Acc w/ MultiWave \\\hline
         MHEALTH & $88.90\% \pm 4.48$ & $96.68\% \pm 2.99$ & $\mathbf{97.59\% \pm 0.68}$ \\\hline
    \end{tabular}
    \caption{The AUC results with and without the inclusion of frequency masks. We used the CNN-Attn model as a component for the WESAD dataset and the Transformer model for the COVID-19 dataset. We also report the accuracy results for MHEALTH dataset with the LSTM model. We selected these models for the ablation study because MultiWave brought the greatest improvement to the performance of the baseline models, which made it easier to reliably quantify the benefit brought by the frequency splitting and the benefit brought by the sparsity-inducing masks.}
    \label{tab:NoMask}
\end{table*}

\subsection{The effect of frequency masks}
The outcomes displayed in Table~\ref{tab:NoMask} show how incorporating frequency masks impacts the performance of MultiWave on the three datasets. The inclusion of component masks enhances MultiWave's performance in all cases, indicating that the use of frequency masks leads to improved performance by eliminating unnecessary frequency components from the signals.
\section{Conclusions}
In this paper, we presented a new framework called MultiWave that augments any deep learning time series model with components that operate at different frequencies of signals using wavelet decomposition. We further improved MultiWave by introducing frequency masks, which remove noninformative frequency components of signals from the component inputs. We show that this framework improves the performance of time series models in synthetic datasets, as well as three real-world datasets for wearable affect and stress detection, COVID-19 prediction, and human activity recognition.
\clearpage

\bibliography{references}

\appendix

\clearpage
\section{WESAD}
\label{sec:app_WESAD}
\subsection{Preprocessing}
\label{sec:app_WESAD_preprocess}
We follow \cite{Ditch} to preprocess the data and use a similar technique to down-sample the signals. However, instead of the frequencies used in the article, we used frequencies that are powers of 2 given in Table~\ref{tab:WESAD_freqs}. We used a five-fold data splitting technique to train our model. For each fold, we trained the model on three folds, validated it on one fold, and tested it on the remaining fold. We do this five times with random seed values of 123 - 127 for all the models, and the mean and standard deviation are reported in the paper.
\subsection{Hyperparameters}
We perform a grid search on the validation data to select the hyperparameters for the baseline model. Then, we set the sizes of the MultiWave components so that they have a similar number of parameters as the baseline model. The hyperparameters selected for the WESAD dataset are given in Table~\ref{tab:WESAD_hyp}. 

\section{COVID-19 dataset}
\label{sec:app_COVID}
\subsection{Preprocessing}
This is an irregularly sampled dataset, we resample features based on their overall rate in all training data according to Table~\ref{tab:COVID_freqs}. If a feature is missing at the time of sampling, we used the closest available value. If a feature is completely missing for a patient, we fill the values with the average of that feature among all the training samples. We also normalize the features to values between 0 and 1 using a min-max scaler.

\subsection{Hyperparameters}
We use the validation data to select the hyperparameters for the baseline model using grid search, and then for MultiWave components, we set the model sizes such that the number of parameters is approximately the same between the two models. We use data from 50 subjects for validation. The hyperparameters selected for MultiWave in this dataset are given in Table~\ref{tab:COVID_hyp}.
\section{MHEALTH}
\label{sec:app_Mhealth}
\subsection{Preprocessing}
To prepare the dataset for analysis, several steps were taken. First, we resampled the data for the ``no activity" category to 30000, bringing it in line with the other activities. Our experiment was designed to use only accelerometer and gyroscope data, which is more challenging and better simulates real-world conditions. We then eliminated duplicate values and removed any outliers falling below the 0.01 quantile or above the 0.99 quantile for each feature. Finally, we selected windows of size 100 and retained the original activity labels. To split the dataset into training and testing, we utilized data from subjects 9 and 10 for testing and the remaining subjects for training.
\subsection{Hyperparameters}
Similar to the previous datasets, we used the validation data to perform a grid search and select the hyperparameters for the baseline model. As for the MultiWave components, we adjust the model sizes to ensure that they have roughly the same number of parameters as the baseline model. The hyperparameters selected for the MHEALTH dataset are given in Table~\ref{tab:MHEALTH_hyp}.

\section{\rebuttal{Experiments with frequency components that are not powers of 2}}
\label{sec:challenging_Syn}
\begin{figure}[h!]
	\begin{center}
        \includegraphics[width=0.45\textwidth]{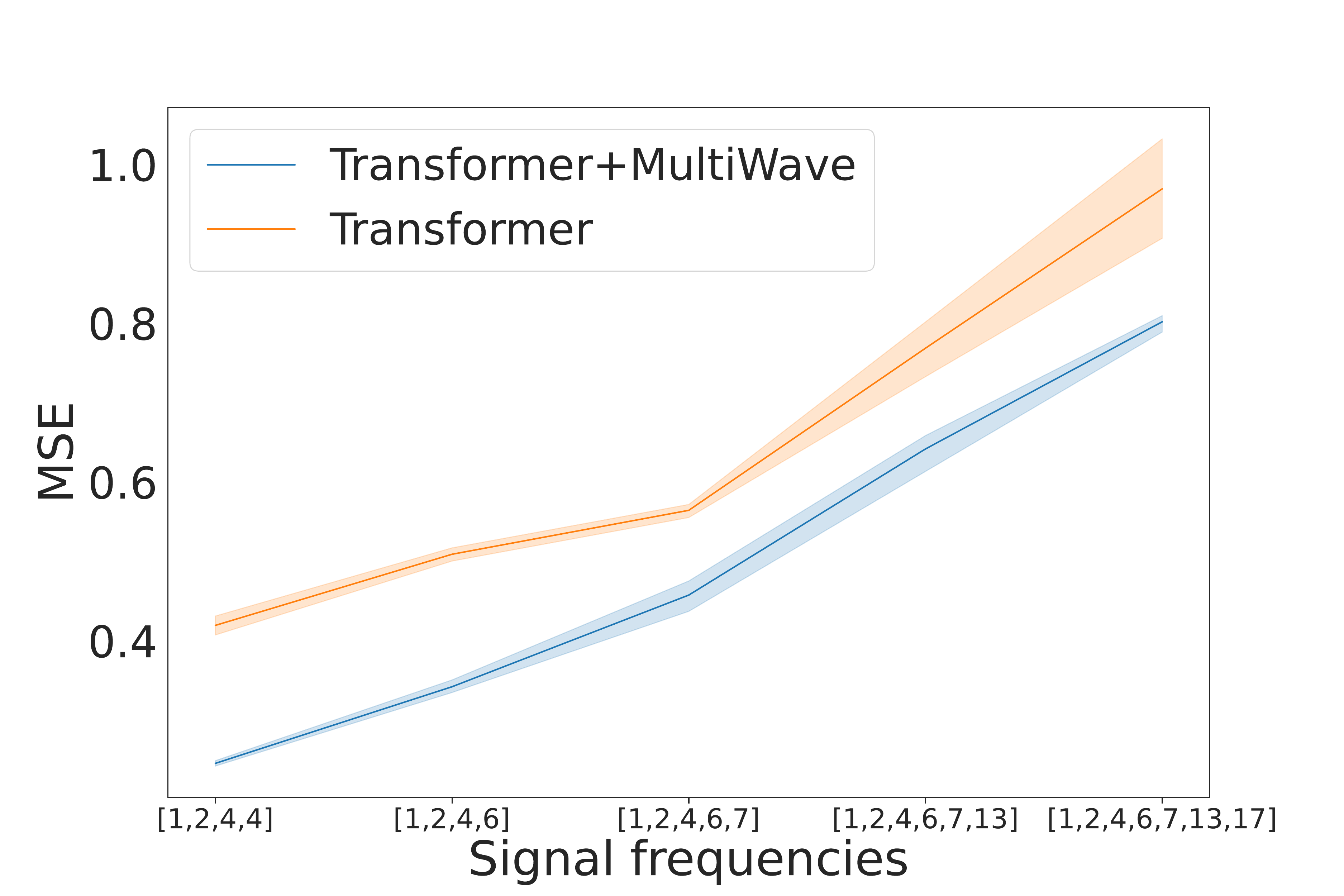}
		\caption{\rebuttal{The Mean Squared Error (MSE) results for a synthetic dataset, on Transformer model with and without the addition of MultiWave. We add signals with frequencies 6Hz, 7Hz, 13Hz and 17Hz.}}
		\label{fig:SynResults_nonpower2}
	\end{center}
\end{figure}
\rebuttal{For this experiment, we repeated experiment 1 from section~\ref{sec:syn_data} using the Transformer model with varying frequencies, but we added signals with frequencies of 6 Hz, 7 Hz, 13 Hz, and 17 Hz, instead of using powers of 2. We began with signals having ${1Hz, 2Hz, 4Hz, 4Hz}$ frequencies and changed the frequency of the last signal from 4 to 6 to observe how the models' performance would be affected, before adding the remaining signals. The results are displayed in figure~\ref{fig:SynResults_nonpower2}, which demonstrates that despite the added signal frequencies not being powers of 2, MultiWave consistently improves performance.}

\section{\rebuttal{Model with FFT features}}
\label{sec:ModelFFT}
\begin{figure}[h!]
    \centering
    \includegraphics[width=0.4\textwidth]{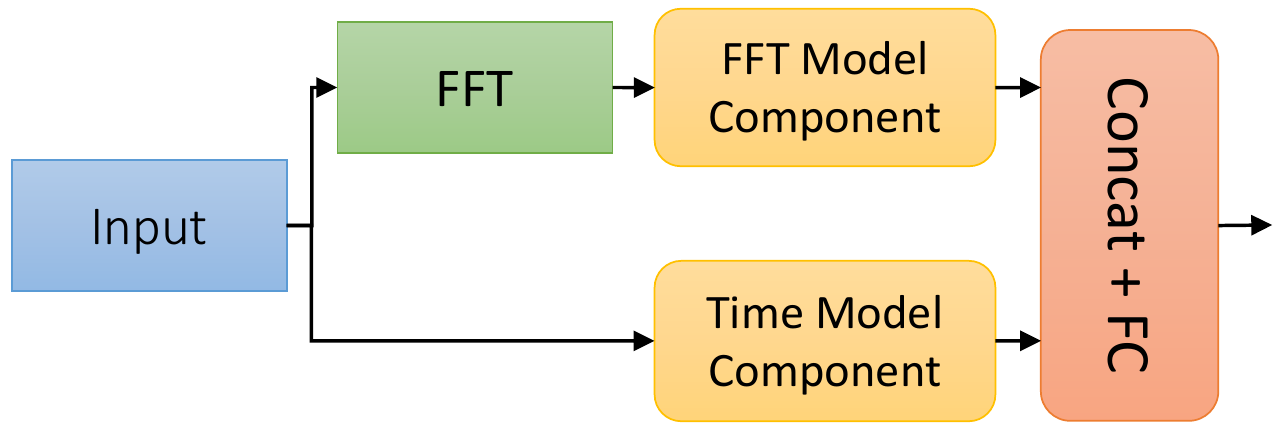}
    \caption{Model with FFT features included}
    \label{fig:FFTModel}
\end{figure}

\rebuttal{To investigate whether the performance improvements of the models are solely attributable to the inclusion of frequency features, we constructed a naive model that utilizes both FFT features and the signal itself. The architecture of this model is illustrated in Figure~\ref{fig:FFTModel}. We used the same model component for both FFT features and the input signal, and concatenated the results before passing them through a 2-layer fully connected network. The experimental results are presented in Table~\ref{tab:WESADFFT}.}

\begin{table*}
    \centering
    \begin{tabular}{|c|c|c|}
        \hline
        Signal & Original Sampling & Downsampled to \\ \hline
        ECG RespiBAN & $700$ Hz & $64$ Hz \\ \hline
        ACC RespiBAN & $700$ Hz & $8$ Hz \\ \hline
        EMG RespiBAN & $700$ Hz & $8$ Hz \\ \hline
        EDA RespiBAN & $700$ Hz & $4$ Hz \\ \hline
        TEMP RespiBAN & $700$ Hz & $4$ Hz \\ \hline
        Respiration RespiBAN & $700$ Hz & $4$ Hz \\ \hline
        BVP Empatica & $64$ Hz & $64$ Hz \\ \hline
        ACC Empatica & $32$ Hz & $8$ Hz \\ \hline
        EDA Empatica & $4$ Hz & $4$ Hz \\ \hline
        TEMP Empatica & $4$ Hz & $4$ Hz \\ \hline
    \end{tabular}
    \caption{WESAD dataset feature frequencies}
    \label{tab:WESAD_freqs}
\end{table*}

\begin{table*}
    \centering
    \begin{tabular}{|p{13cm}|p{1.2cm}|}
        \hline
        Signals & Sampled Rates \\ \hline
    hemoglobin, Serum chloride, Prothrombin time, procalcitonin,  eosinophils(\%), Alkaline phosphatase, albumin, basophil(\%), Total bilirubin, Platelet count, monocytes(\%), indirect bilirubin, Red blood cell distribution width, neutrophils(\%), total protein, Prothrombin activity, mean corpuscular volume, hematocrit, White blood cell count, mean corpuscular hemoglobin concentration, fibrinogen, Urea, lymphocyte count, Red blood cell count, Eosinophil count, Corrected calcium, Serum potassium, glucose, neutrophils count, Direct bilirubin, Mean platelet volume, RBC distribution width SD, Thrombin time, (\%)lymphocyte, D-D dimer, Total cholesterol, aspartate aminotransferase, Uric acid, HCO3-, calcium, Lactate dehydrogenase, platelet large cell ratio, monocytes count, PLT distribution width, globulin, $\gamma$-glutamyl transpeptidase, International standard ratio, basophil count(\#), mean corpuscular hemoglobin, Activation of partial thromboplastin time, High sensitivity C-reactive protein, serum sodium, thrombocytocrit, glutamic-pyruvic transaminase,eGFR, creatinine & every day \\\hline
    antithrombin, Quantification of Treponema pallidum antibodies, HBsAg, HCV antibody quantification, Amino-terminal brain natriuretic peptide precursor(NT-proBNP), Fibrin degradation products, HIV antibody quantification, ESR & every 2 days \\ \hline
    PH value & every 3 days \\ \hline
    Interleukin 2 receptor, Interleukin 10, Interleukin 8, Tumor necrosis factor$\alpha$, Interleukin 1$\beta$, Interleukin 6 & every 5 days \\ \hline
    2019-nCoV nucleic acid detection & every 7 days \\ \hline
    ferritin & every 8 days \\ \hline
    \end{tabular}
    \caption{COVID dataset feature frequencies}
    \label{tab:COVID_freqs}
\end{table*}

\begin{table}
    \centering
    \begin{tabular}{|c|c|c|}
        \hline
        Model & Hyper parameter & Value \\ \hline\hline
        {\multirow{9}{*}{LSTM}} & LSTM cell size & 28 \\ \cline{2-3}
                                 & Initial Mask Weight values& $0.5$\\ \cline{2-3}
                                 & Mask norm weight & $0.05$ \\ \cline{2-3}
                                 & Number of Layers & $1$ \\ \cline{2-3}
                                 & Learning rate & $0.001$ \\ \cline{2-3}
                                 & Batch size & $16$ \\ \cline{2-3}
                                 & patience & $15$ \\ \cline{2-3}
                                 & Number of Components & $6$ \\ \cline{2-3}
                                 & AddBaseline & True \\ \hline
        {\multirow{9}{*}{CNN-Attn}} & CNN Kernel size & 7 \\ \cline{2-3}
                                 & Initial Mask Weight values& $0.5$\\ \cline{2-3}
                                 & Mask norm weight & $0.1$ \\ \cline{2-3}
                                 & Number of Layers & $2$ \\ \cline{2-3}
                                 & Learning rate & $0.001$ \\ \cline{2-3}
                                 & Batch size & $16$ \\ \cline{2-3}
                                 & patience & $15$ \\ \cline{2-3}
                                 & Number of Components & $6$ \\ \cline{2-3}
                                 & AddBaseline & True \\ \hline
        {\multirow{9}{*}{CNN-LSTM}} & CNN Kernel size & 7 \\ \cline{2-3}
                                 & Initial Mask Weight values& $0.5$\\ \cline{2-3}
                                 & Mask norm weight & $0.1$ \\ \cline{2-3}
                                 & Number of Layers & $2$ \\ \cline{2-3}
                                 & Learning rate & $0.001$ \\ \cline{2-3}
                                 & Batch size & $16$ \\ \cline{2-3}
                                 & patience & $15$ \\ \cline{2-3}
                                 & Number of Components & $6$ \\ \cline{2-3}
                                 & AddBaseline & False \\ \hline
        {\multirow{9}{*}{FCN}} & CNN Kernel size & 7 \\ \cline{2-3}
                                 & Initial Mask Weight values& $0.5$\\ \cline{2-3}
                                 & Mask norm weight & $0.1$ \\ \cline{2-3}
                                 & Number of Layers & $2$ \\ \cline{2-3}
                                 & Learning rate & $0.001$ \\ \cline{2-3}
                                 & Batch size & $16$ \\ \cline{2-3}
                                 & patience & $15$ \\ \cline{2-3}
                                 & Number of Components & $6$ \\ \cline{2-3}
                                 & AddBaseline & False \\ \hline
    \end{tabular}
    \caption{The selected MultiWave hyperparameters for WESAD dataset}
    \label{tab:WESAD_hyp}
\end{table}

\begin{table}
    \centering
    \begin{tabular}{|c|c|c|}
        \hline
        Model & Hyper parameter & Value \\ \hline\hline
        {\multirow{9}{*}{LSTM}} & LSTM cell size & 19 \\ \cline{2-3}
                                 & Initial Mask Weight values& $0.3$\\ \cline{2-3}
                                 & Mask norm weight & $0.1$ \\ \cline{2-3}
                                 & Number of Layers & $1$ \\ \cline{2-3}
                                 & Learning rate & $0.001$ \\ \cline{2-3}
                                 & Batch size & $16$ \\ \cline{2-3}
                                 & patience & $25$ \\ \cline{2-3}
                                 & Number of Components & $6$ \\ \cline{2-3}
                                 & Reset Model & False \\ \hline
        {\multirow{9}{*}{CNN-Attn}} & CNN Kernel size & 7 \\ \cline{2-3}
                                 & Initial Mask Weight values& $0.3$\\ \cline{2-3}
                                 & Mask norm weight & $0.1$ \\ \cline{2-3}
                                 & Number of Layers & $2$ \\ \cline{2-3}
                                 & Learning rate & $0.001$ \\ \cline{2-3}
                                 & Batch size & $16$ \\ \cline{2-3}
                                 & patience & $25$ \\ \cline{2-3}
                                 & Number of Components & $6$ \\ \cline{2-3}
                                 & Reset Model & False \\ \hline
        {\multirow{9}{*}{CNN-LSTM}} & CNN Kernel size & 7 \\ \cline{2-3}
                                 & Initial Mask Weight values& $0.3$\\ \cline{2-3}
                                 & Mask norm weight & $0.1$ \\ \cline{2-3}
                                 & Number of Layers & $2$ \\ \cline{2-3}
                                 & Learning rate & $0.001$ \\ \cline{2-3}
                                 & Batch size & $16$ \\ \cline{2-3}
                                 & patience & $25$ \\ \cline{2-3}
                                 & Number of Components & $6$ \\ \cline{2-3}
                                 & Reset Model & False \\ \hline
        {\multirow{9}{*}{Transformer}} & Hidden Embedding Size & 13 \\ \cline{2-3}
                                 & Initial Mask Weight values& $0.3$\\ \cline{2-3}
                                 & Mask norm weight & $0.1$ \\ \cline{2-3}
                                 & Number of Layers & $4$ \\ \cline{2-3}
                                 & Number of Heads & $3$ \\ \cline{2-3}
                                 & Learning rate & $0.001$ \\ \cline{2-3}
                                 & Batch size & $16$ \\ \cline{2-3}
                                 & patience & $25$ \\ \cline{2-3}
                                 & Reset Model & False \\ \hline
    \end{tabular}
    \caption{The selected MultiWave hyperparameters for COVID dataset}
    \label{tab:COVID_hyp}
\end{table}

\begin{table}
    \centering
    \begin{tabular}{|c|c|c|}
        \hline
        Model & Hyper parameter & Value \\ \hline\hline
        {\multirow{9}{*}{LSTM}} & LSTM cell size & 16 \\ \cline{2-3}
                                 & Initial Mask Weight values& $1.5$\\ \cline{2-3}
                                 & Mask norm weight & $0.001$ \\ \cline{2-3}
                                 & Number of Layers & $1$ \\ \cline{2-3}
                                 & Learning rate & $0.001$ \\ \cline{2-3}
                                 & Batch size & $32$ \\ \cline{2-3}
                                 & patience & $15$ \\ \cline{2-3}
                                 & Number of Components & $6$ \\ \cline{2-3}
                                 & Reset Model & False \\ \hline
        {\multirow{9}{*}{CNN-Attn}} & CNN Kernel size & 7 \\ \cline{2-3}
                                 & Initial Mask Weight values& $0.5$\\ \cline{2-3}
                                 & Mask norm weight & $0.01$ \\ \cline{2-3}
                                 & Number of Layers & $2$ \\ \cline{2-3}
                                 & Learning rate & $0.001$ \\ \cline{2-3}
                                 & Batch size & $32$ \\ \cline{2-3}
                                 & patience & $15$ \\ \cline{2-3}
                                 & Number of Components & $6$ \\ \cline{2-3}
                                 & Reset Model & False \\ \hline
        {\multirow{9}{*}{CNN-LSTM}} & CNN Kernel size & 7 \\ \cline{2-3}
                                 & Initial Mask Weight values& $1.5$\\ \cline{2-3}
                                 & Mask norm weight & $0.001$ \\ \cline{2-3}
                                 & Number of Layers & $1$ \\ \cline{2-3}
                                 & Learning rate & $0.001$ \\ \cline{2-3}
                                 & Batch size & $32$ \\ \cline{2-3}
                                 & patience & $15$ \\ \cline{2-3}
                                 & Number of Components & $6$ \\ \cline{2-3}
                                 & Reset Model & False \\ \hline
        {\multirow{9}{*}{Transformer}} & Hidden Embedding Size & 12 \\ \cline{2-3}
                                 & Initial Mask Weight values& $0.4$\\ \cline{2-3}
                                 & Mask norm weight & $0.5$ \\ \cline{2-3}
                                 & Number of Layers & $1$ \\ \cline{2-3}
                                 & Number of Heads & $3$ \\ \cline{2-3}
                                 & Learning rate & $0.001$ \\ \cline{2-3}
                                 & Batch size & $32$ \\ \cline{2-3}
                                 & patience & $15$ \\ \cline{2-3}
                                 & Reset Model & False \\ \hline
    \end{tabular}
    \caption{The selected MultiWave hyperparameters for MHEALTH dataset}
    \label{tab:MHEALTH_hyp}
\end{table}

\end{document}